\documentclass[11pt]{article}

\usepackage[T1]{fontenc}
\usepackage[utf8]{inputenc}
\usepackage{lmodern}
\usepackage{amsmath,amssymb,bm}
\usepackage[ruled,vlined]{algorithm2e}
\usepackage{hyperref}
\usepackage{graphicx}
\usepackage{fullpage}
\usepackage[backend=biber,style=numeric-comp,sorting=none]{biblatex}
\usepackage{xcolor}
\usepackage{authblk}
\usepackage{booktabs}
\usepackage{siunitx}

\title{Efficient Per-Example Gradient Computations in\\ Convolutional Neural Networks}
\author[1]{Gaspar Rochette\footnote{Work completed while at Owkin, Inc.}}
\author[2]{Andre Manoel}
\author[2]{Eric W. Tramel}
\affil[1]{{ENS, PSL University, Paris, France}}
\affil[2]{Owkin, Inc. New York, NY}

\date{\today}

\begin{document}
\maketitle

\begin{abstract}
    Deep learning frameworks leverage GPUs to perform massively-parallel
    computations over batches of many training examples efficiently.
    However, for certain tasks, one may be interested in performing
    per-example computations, for instance using per-example gradients to
    evaluate a quantity of interest unique to each example.  One notable
    application comes from the field of differential privacy, where
    per-example gradients must be norm-bounded in order to limit the impact
    of each example on the aggregated batch gradient.  In this work, we
    discuss how per-example gradients can be efficiently computed in
    convolutional neural networks (CNNs). We compare existing strategies by
    performing a few steps of differentially-private training on CNNs of
    varying sizes.  We also introduce a new strategy for per-example
    gradient calculation, which is shown to be advantageous depending on the
    model architecture and how the model is trained. This is a first step in
    making differentially-private training of CNNs practical.
\end{abstract}

\section{Introduction}

Today, as developments have progressed in the field of privacy-aware machine
learning (ML), it has become widely acknowledged that applying ML to
sensitive data introduces a number of privacy risks which must be taken into
consideration when building responsible ML systems.  Multiple works
\cite{shokri2017membership,carlini2018secret,melis2019exploiting,yeom2018privacy}
have shown that some information about training data can be recovered from
the network's parameters, especially in the context of membership inference.
To be able to use sensitive data such as medical records for training
machine learning models, something must be done to ensure the privacy of the data
with respect to the trained model. One approach
to this is \emph{differential privacy} (DP)~\cite{dwork2014algorithmic},
which limits the influence that one specific data point can have on the model
parameters. In the case of ML applied to medical records, the utility in
applying DP would be to provide a mathematical assurance on the level of
potential additional harm that could come to a patient if their records are
included in the training cohort rather than excluded.

A necessary step in the application of DP to the training of ML models via
stochastic gradient descent, as shown in~\cite{abadi2016deep}, is the
application of gradient clipping. Notably, model gradients must be
norm-bounded individually for each sample, not according to the aggregate
across a given mini-batch.  That is, one should consider the gradient of the
loss $\mathcal L_i$ induced by example $i$, rather than the gradient of the
global loss $\sum_i \mathcal L_i$.  This \emph{per-example gradient
clipping} has also been used in the context of reinforcement learning
\cite{wang2016dueling}. In~\cite{abadi2016deep}, gradient clipping is
defined for a given maximum gradient norm bound $C$ by
\begin{equation}
    \bar{g}(x_i) = \frac{g(x_i)}{\max \left(1, \frac{1}{C} ||g(x_i) ||_2 \right)},
    \label{eq:gradclip}
\end{equation}
where $g(x_i) \triangleq \nabla_{\theta} \mathcal{L}_i(\theta, x_i)$ is the
gradient of the model parameters with respect to data input $x_i$. This
gradient clipping has the effect of bounding the maximal amount of influence
a single sample can have on the direction of a model update. When applied over
the course of training, in conjunction with the addition of noise, the
sensitivity of the final model to any one sample is therefore bounded thanks
to the composition properties of DP~\cite{abadi2016deep}.

Unfortunately, auto-differentiation libraries, like those included in
PyTorch and Tensorflow, do not naturally offer the option to compute
per-example gradients as conventional ML training workflows only require
a single gradient produced by averaging across the samples of the
mini-batch. Since a successful implementation of differentially-private
model training requires this per-example gradient clipping, efficiently
producing per-example gradients is a critical step. One approach would be to
start from the bottom and make the necessary changes to existing low-level
auto-differentiation tools to enable this computation.  Another tactic,
which we investigate here, is to start from the high-level framework API (in
this case, PyTorch) and see what options are available to effect this
per-example gradient estimation.

Multiple methods in this vein have been suggested over the past several
years.  One such method was introduced by Goodfellow in a technical
report~\cite{goodfellow2015efficient}.  While very efficient, this method
only applies to linear fully-connected (\emph{dense}) network layers.  In
this report, we explain how to extend the approach
of~\cite{goodfellow2015efficient} to convolutional networks, and proceed to
compare its efficiency to that of other existing approaches.

\section{Existing methods for computing per-example gradients}
\label{sec:clip_methods}

In this section, we review some of the existing techniques for computing
per-example gradients at the level of the of the ML framework API.
Algorithmic optimization can accelerate the computation, but the main
speed-ups come from fully utilizing the parallelization of operations on
GPU. The three main approaches to per-example gradient computation are
described as follows:

\paragraph{Naive approach.}
The simplest strategy consists in changing the batch size to $1$. Doing so,
one can iterate over the originally desired batch size, computing the
gradient corresponding to each sample in the batch. Since there is no
parallelization, this method is very slow. This approach requires a minimal
amount of code. We denote it by \texttt{naive} in our experiments.

\paragraph{Changing one step of the backpropagation.}
In~\cite{goodfellow2015efficient}, it was shown that one can use the
auto-differentiation's intermediate results to compute the per-example gradient
by hand. However, this formulation does not allow for convolutions
to be used efficiently. In practice, when attempting to apply DP to training
Deep Convolutional Neural Networks (DCNN) models, multiple papers have used
a transfer learning approach, training only the last linear task layers
while freezing a set of pre-trained CNN feature extraction layers
\cite{abadi2016deep,alain2015variance,bhowmick2018protection}. We explain in
the next section how to extend this computation to CNNs using functions
available in PyTorch. We denote it by \texttt{crb} in our experiments, for
\emph{chain rule based}.

\paragraph{Using multiple copies of the model.}
This strategy relies on the same idea as the first one: using mini-batches
of size 1. In this approach, however, one copies the model as many times as there
are samples in the batch. Each model is used on only one example, thus
parallelizing the iteration from the first approach. This approach is very
fast and requires only a small amount of code.
In addition to this, the model duplicates can share their parameters with
the original model in order to minimize the training memory footprint.
This method has been proposed by Goodfellow in 2017, on a Github
thread\footnote{See discussion at
\url{https://github.com/tensorflow/tensorflow/issues/4897\#issuecomment-290997283}.}.
We denote it by \texttt{multi} in our experiments.

\section{Per-example backpropagation for convolutional networks}

In this section, we describe a way to compute per-example gradients from
partial derivatives explicitly computed by auto-differentiation tools. We
only describe the computation of these gradients for a single linear layer, but
the extension to multiple non-linear layers is straightforward. Doing so
for each layer in the network results in full per-example gradients.
This method is denoted \texttt{crb} in our experiments, which stands for
\emph{chain rule based}. We provide an implementation of this
method at \url{https://github.com/owkin/grad-cnns}.

\subsection{Fully-connected linear layers}
In the case of dense linear layers, one can refer directly to the 
technique introduced in~\cite{goodfellow2015efficient}, which we
reproduce, here.
Let $H$ be a linear layer of a neural network, with input $x = \left[ x_1,
\hdots, x_I \right]^T$ and output $y = Hx = \left[ y_1, \hdots, y_J
\right]^T$. When training a model, we wish to update the model parameters
$H$ so as to minimize some output-depending loss function $\mathcal{L}$, i.e.
following the negative gradient of $\mathcal{L}$ w.r.t. $H$.
This gradient is obtained from the model output via backpropagation,
\begin{equation}
    \frac{\partial \mathcal L}{\partial H_{kl}}
    = \sum_j \frac{\partial y_j}{\partial H_{kl}} \,\frac{\partial \mathcal L}{\partial y_j}
    = x_l \, \frac{\partial \mathcal L}{\partial y_k} \, .
\end{equation}
The \emph{trick} of calculating per-example gradients post auto-differentiation
comes from the observation of the final equality above: the partial w.r.t. a single coefficient
of $H$ comes from the simple multiplication of a single layer input with the partial
of the loss with respect to a single output. 
As noted by \cite{goodfellow2015efficient}, one can store the layer's input
$x$, and auto-differentiation tools give access to $\frac{\partial \mathcal
L}{\partial y}$ for each sample in the batch. We can then compute the
gradient with simple batch matrix multiplications, an outer-product computation,
which can be done efficiently on GPU, $\nabla_H\mathcal{L} = (\nabla_y\mathcal{L})\, x^T$.

\subsection{Application to convolutional layers}
The computation described above uses linear matrix representation, which
if applied to the case of convolution would be extremely inefficient. We describe in this
section how to use convolution operations to obtain the same result when
back-propagating through convolutional layers. 
We will do so using PyTorch format for tensors
dimensions, i.e. $\texttt{(batch, channels, spatial dimensions)}$.

\subsubsection{Notation}

\paragraph*{Spatial dimensions.}
In order not to overcomplicate the derivation, we will work with 1D
convolutions. We will refer to this spatial dimension with the variable $t
\in \{0 \hdots T - 1\}$, or in the case of the convolutional kernel, with the
letter $k \in \{0 \hdots K - 1\}$. Note that the same derivations are correct
for other dimensions.

\paragraph*{Channels.}
Channels will be an essential part of our derivation, especially when
dealing with the \verb?groups? argument. We will refer to the input and
output channels respectively with the variables $c \in \{0 \hdots C -1\}$ and
$d \in \{0 \hdots D - 1\}$.

\paragraph*{Batch.}
We will refer to the batch dimension with the variable $b \in \{0 \hdots B - 1\}$.

\subsubsection{Standard convolution}
\label{sec:convolution}

We start with standard convolution as implemented in most ML frameworks;
a set of filters, or \emph{kernels}, $h$ of shape $(D, C, K)$
is convolved with the input tensor $x$ of shape $(B, C, T)$ to form an output
tensor $y$ of shape $(B, D, T-K+1)$. This discrete convolution is given by the
formula\footnote{Note
that our definition of convolution uses an offset $k$ rather than $-k$. This
is both for mathematical simplicity and for consistency with PyTorch, which
uses this convention as well.}
\begin{equation}
    \label{eq:convolution}
    y[b, d, t] = (x \ast h)[b, d, t] \triangleq \sum_{c=0}^{C-1} \sum_{k=0}^{K-1} x[b, c, t+k] \, h[d, c, k] \, ,
\end{equation}
where we note the aggregation step performed over the $C$ input channels to
produce the final convolved output of the layer.
To calculate per-example gradients with respect to $h$,  
we are interested in differentiating sample $b$'s contribution to the loss
$\mathcal L$, that is, $\mathcal L[b]$ where $\mathcal L = \sum_b \mathcal
L[b]$. By applying the same technique as in the fully-connected case, but carrying
through the convolution operation, we see that the gradient with respect to the convolution kernel is
\begin{align*}
    \frac{\partial \mathcal L[b]}{\partial h[d, c, k]}
    &= \sum_{t=0}^{T-K} \frac{\partial y[b, d, t]}{\partial h[d, c, k]}
        \, \frac{\partial \mathcal L[b]}{\partial y[b, d, t]} \\
    &= \sum_{t=0}^{T-K} \frac{\partial \left( \sum_{\tilde c=0}^{C-1}
        \sum_{\tilde k=0}^{K-1} x[b, \tilde c, t+\tilde k] \, h[d, \tilde c,
        \tilde k] \right)}{\partial h[d, c, k]} \, \frac{\partial \mathcal
        L[b]}{\partial y[b, d, t]} \\
    &= \sum_{t=0}^{T-K} x[b, c, t+k] \, \underbrace{\frac{\partial \mathcal
        L[b]}{\partial y[b, d, t]}}_{\nabla_y \mathcal{L} [b, d, t]} \, .
\end{align*}
Since the spatial dimensions of $x$ and $\nabla_y \mathcal{L}$ are
$T$ and $T-K+1$ respectively, their convolution has the same spatial dimension $K$ as the
convolution kernel. This convolution can be seen as a \emph{per-example
convolution}, which we denote by $\circledast$:
\begin{equation}
    \frac{\partial \mathcal L[b]}{\partial h[d, c, k]} = \left( x
    \circledast \nabla_y \mathcal{L} \right)[b, d, c, k] \, \triangleq
    \sum_{t=0}^{T-K} x[b, c, t+k] \, \nabla_y \mathcal{L} [b, d, t].
    \label{eq:formula}
\end{equation}
This convolution operation, although similar to the one in~\eqref{eq:convolution}, 
is not available in auto-differentiation libraries.
However, the \texttt{groups} argument in PyTorch's regular convolution
allows one to evaluate such an operation. Recall that \texttt{groups} splits
the input tensor into groups of the same size, performs independent convolutions
on each group, and finally concatenates the resulting outputs. More
precisely, one can reshape the input $x$ to size $(B, C/G, G, T)$, and the kernel
$h$ to $(D/G, G, C/G, K)$ and subsequently define the group convolution as
\begin{equation}
    \forall d \in \{0 \hdots D/G-1\} ,\quad y[b, g\frac{D}{G} + d, t] = (x
    \ast_G h) [b, d, g, t] \triangleq \sum_{c=0}^{C/G-1} \sum_{k=0}^{K-1}
    x[b, c, g\frac{C}{G} + c, t+k] \, h[d, g, c, k].
    \label{eq:groupconv}
\end{equation}
In Algorithm \ref{algo:batchconv}, we explain how to use the group
convolution (\ref{eq:groupconv}) in order to evaluate (\ref{eq:formula}).
We first replace $h$ by $\nabla_y \mathcal{L}$ in the formula above, and 
swap $t$ and $k$, also making sure that $K$ is set to $T - K + 1$.
Axes are then swapped and/or combined in such a way that (\ref{eq:formula})
is recovered. Namely, $x$ is reshaped to $(1, B, C, T)$, so that the
effective batch size is 1 and the number of input channels is $B$; and
$\nabla_y \mathcal{L}$ is reshaped to $(BD, 1, 1, T - K + 1)$, so that the
number of output channels is $BD$. The key is in noting that, since each sample
is treated as a different input channel, grouping allows us to treat each of
them in parallel. Finally, we note that after reshaping the inputs to the
convolution, it is necessary, even though we operate on a 1D spatial
dimension, to utilize a 2D convolution.  Analogously, when adapting this
procedure to the estimation of per-example gradients for 2D convolutional
layers, it is necessary to use a 3D convolution.

\begin{figure}[ht]
    \begin{algorithm}[H]
        \vskip .1cm
        \KwIn{$x$ of size $(B, C, T)$, $\delta y\triangleq \nabla_y\mathcal{L}$ of size $(B, D, T-K+1)$}
        \KwResult{$\delta h \triangleq \nabla_h\mathcal{L}$ of size $(B, D, C, K)$}

        \# Reshape $x$ and $\delta y$\\
        \texttt{$x$ $\leftarrow$ reshape $x$ to shape $(1, B, C, T)$} \\
        \texttt{$\delta y$ $\leftarrow$ reshape $\delta y$ to shape $(B \times D, 1, 1, T-K+1)$}\\
        \vskip .3cm

        \# Call regular convolution function, with one extra dimension \\
        \texttt{$\delta y$ $\leftarrow$ conv2d($x$, $\delta y$, groups=$B$)} \hskip .5cm \# shape $(1, B \times D, C, K)$\\
        \vskip .3cm
        \# Reshape output\\
        \texttt{$\delta h$ $\leftarrow$ reshape $\delta h$ to shape $(B, D, C, K)$}\\
        \texttt{return $\delta h$}\\
        \vskip .2cm

        \label{algo:batchconv}
        \caption{Standard convolution: per-example gradients.}
    \end{algorithm}
\end{figure}

\begin{figure}[ht]
    \begin{algorithm}[H]
        \vskip .1cm
        \KwIn{$x$ of size $(B, C, T)$, $\delta y\triangleq\nabla_y\mathcal{L}$ of size $(B, D, T')$, number of groups $\Gamma$, padding size $\Pi$, stride $\Sigma$ and dilation $\Delta$
        }
        \KwResult{$\delta h\triangleq\nabla_h\mathcal{L}$ of size $(B, D, C, K)$}
        \# Reshape $x$ and $y$\\
        \texttt{$x$ $\leftarrow$ reshape $x$ to shape $(1, B \times \Gamma, C/\Gamma, T)$} \\
        \texttt{$\delta y$ $\leftarrow$ reshape $\delta y$ to shape $(B \times D, 1, 1, T')$}\\
        \vskip .3cm
        
        \# Define arguments for convolution with one extra spatial dimension \\
        \texttt{$\Gamma' \leftarrow B \times \Gamma$}\\
        \texttt{$\Pi' \leftarrow (0, \Pi)$}  \hskip .5cm \# zero padding on extra dimension \\
        \texttt{$\Sigma' \leftarrow (1, \Delta)$}  \hskip .5cm \# arguments $\Sigma$ and $\Delta$ are switched \\
        \texttt{$\Delta' \leftarrow (1, \Sigma)$}  \hskip .5cm \# both arguments are one on extra dimension \\
        
        \vskip .3cm
        \# Call convolution function, with one extra dimension \\
        \texttt{$\delta h \leftarrow$ conv2d($x$, $\delta y$, groups=$\Gamma'$, padding=$\Pi'$, stride=$\Sigma'$, dilation=$\Delta'$)}\\
        
        \vskip .3cm
        \# Convolution output shape $(1, B \times D, C/\Gamma, ?)$ must be
        truncated to expected size \\
        \texttt{$\delta h \leftarrow \delta h$[..., :$K$]}
        
        \vskip .3cm
        \# Reshape output \\
        \texttt{$\delta h \leftarrow$ reshape $\delta h$ to shape $(B, D, C/\Gamma, K)$}\\
        \texttt{return $\delta h$}\\
        \vskip .2cm

        \label{algo:batchconvcomplete}
        \caption{Per-example gradients for convolution layer with arguments}
    \end{algorithm}
\end{figure}

\subsubsection{More options for convolutional layers}
In the previous section we described how to compute per-example gradients for
the case of a simple convolution. However, often one wants to control the 
stride and dilation of a convolutional layer, as well padding and grouping. 
We now describe how to integrate into the algorithm all arguments for
convolutions available in PyTorch. The resulting algorithm is described in
Algorithm \ref{algo:batchconvcomplete}.

\paragraph{Stride and Dilation.}
First, let us recall the definition of convolutions with stride or dilation arguments.\\
\begin{itemize}
\item \emph{Convolution with stride $s$}.  The output $y$ has dimension $(B, D, \lfloor \frac{T-K+1}{s} \rfloor)$,
\begin{equation*}
        y[b, d, t] = \sum_{c=0}^{C-1} \sum_{k=0}^{K-1} x[b, c, s t + k] \, h[d, c, k] \, .
    \end{equation*}
\item \emph{Convolution with dilation $r$}. The output $y$ has dimension $(B, D, T - r (K-1))$,
    \begin{equation*}
        y[b, d, t] = \sum_{c=0}^{C-1} \sum_{k=0}^{K-1} x[b, c, t + r k] \, h[d, c, k] \, .
    \end{equation*}
\end{itemize}
Note the only difference in the definition of the two operations is only on to which
time index the scaling multiplier is applied.
Because the roles of $t$ and $k$ are essentially switched in \eqref{eq:formula}, 
it suffices to switch those arguments in the convolution.
That is, if the convolutional layer uses stride $s$ and dilation $r$, the
convolution in~\eqref{eq:formula} should have stride $r$ and dilation
$s$.

Finally, because of the floor operation involved in the output's
size with strided convolution, the output of~\eqref{eq:formula} may have more
output dimensions
than the convolution kernel itself. If this is the case, the extra values
should be ignored, e.g.
\begin{equation*}
    \nabla_h \mathcal{L}
    = \left(x \circledast \nabla_y\mathcal{L}\right)
    [:,\, :,\, :,\, 0:(K-1)] \, ,
\end{equation*}
using Python slicing notation.

\paragraph{Padding and Groups.}
The \verb?padding? and \verb?groups? arguments from the convolutional layer
can be reused in ~\eqref{eq:formula}. For PyTorch, this means using the
options
\begin{itemize}
    \item \texttt{padding = layer.padding},
    \item \texttt{groups = batch\_size * layer.groups},
\end{itemize}
when defining the convolution. Here, the \texttt{layer} variable refers 
to the layer on which we are currently attempting to calculate the per-example
gradients. Note that the value of \texttt{groups} needs to be modified with
the input batch size, after the reshaping of $x$.

\section{Benchmarks}

\begin{figure}[ht]
    \centering
    \includegraphics[width=\textwidth]{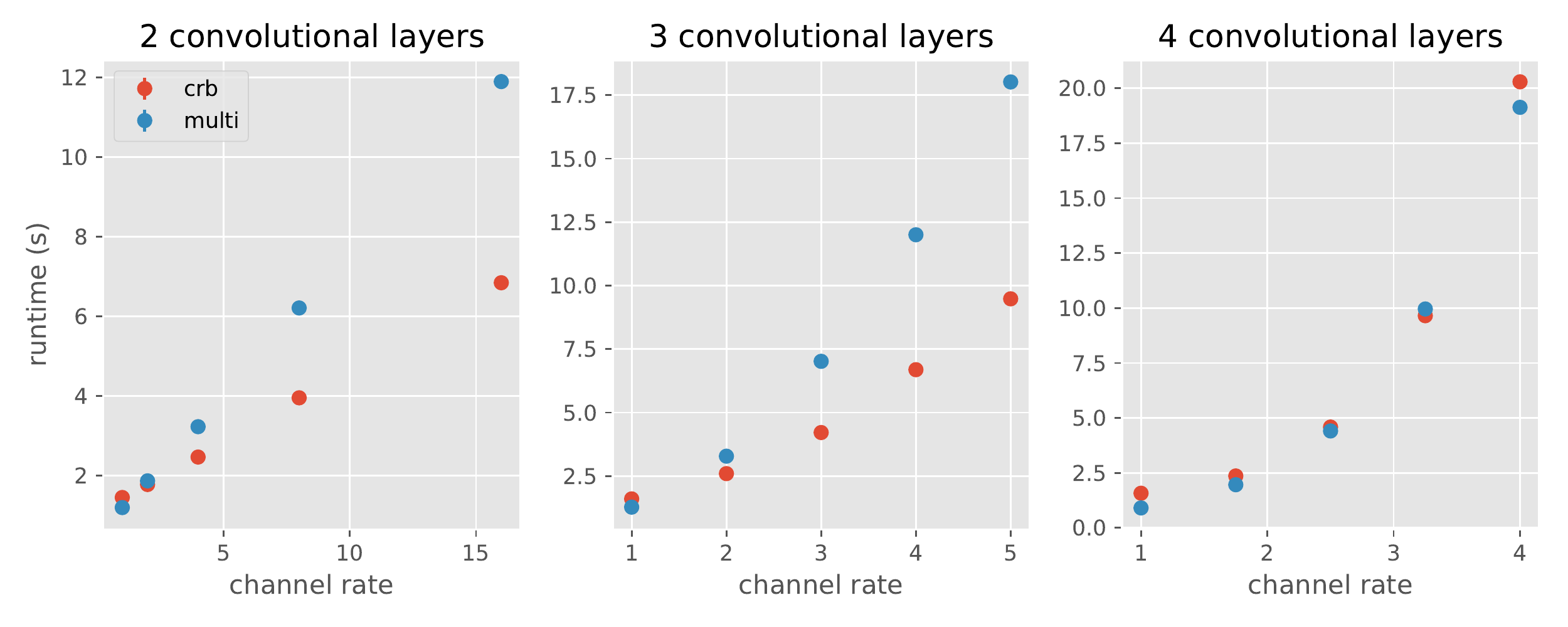}
    \caption{Runtime in seconds for processing 20 batches of 8 examples, in
    CNNs of increasing size. The channel rate is defined as the ratio
    between the number of channels from a layer to the previous, considering
    the first layer has 25 channels. Convolutions are performed with a kernel
    of size 3; ReLU activations are added after each convolution, and a
    max-pooling layer is added after every 2 convolutional layers. Inputs
    are randomly generated and have size $3 \times 256 \times 256$. Left,
    center and right show results for 2, 3 and 4 layers respectively. Each
    point is the average over 10 runs.}
    \label{fig:experiment_cnns}
\end{figure}

As explained above, the \texttt{multi} approach consists in
\emph{vectorizing} the naive approach by creating multiple copies of the same
model and backpropagating through all batch samples in parallel. One
could criticize this approach for using a very large amount of
memory, as many copies need to be created. However, it turns out this can
be done without a single copy of the original model: this can be done using
only pointers to the original parameters, which uses as much memory as other
methods.

That being said, it is not clear which strategy will perform better in terms
of runtime. Due to how computations are optimized in the GPU, analyzing the
computational complexity of each approach is not necessarily useful. We will
thus run a series of experiments in order to perform this comparison
empirically.

All the experiments in this section have been performed on \texttt{n1-standard-8}
instance on GCP, with a Nvidia P100 GPU. The image used is based on Ubuntu 18.04,
and had Python 3.5.3 and PyTorch 1.1 installed.

\subsection{Toy networks}

In a first experiment, we create convolutional architectures with 2, 3 and 4
sequential convolutional layers such that the number of channels from a layer to the next increases
according to a given ratio. In Fig.~\ref{fig:experiment_cnns}, we show
that, for shallower networks, \texttt{crb} runs faster than \texttt{multi}
as we increase this ratio. Increasing the number of layers, however, 
seems to be an advantage for \texttt{multi}. Finally, in a network with 4
layers, the two methods are competitive.

\begin{figure}[ht]
    \centering
    \includegraphics[width=0.4\textwidth]{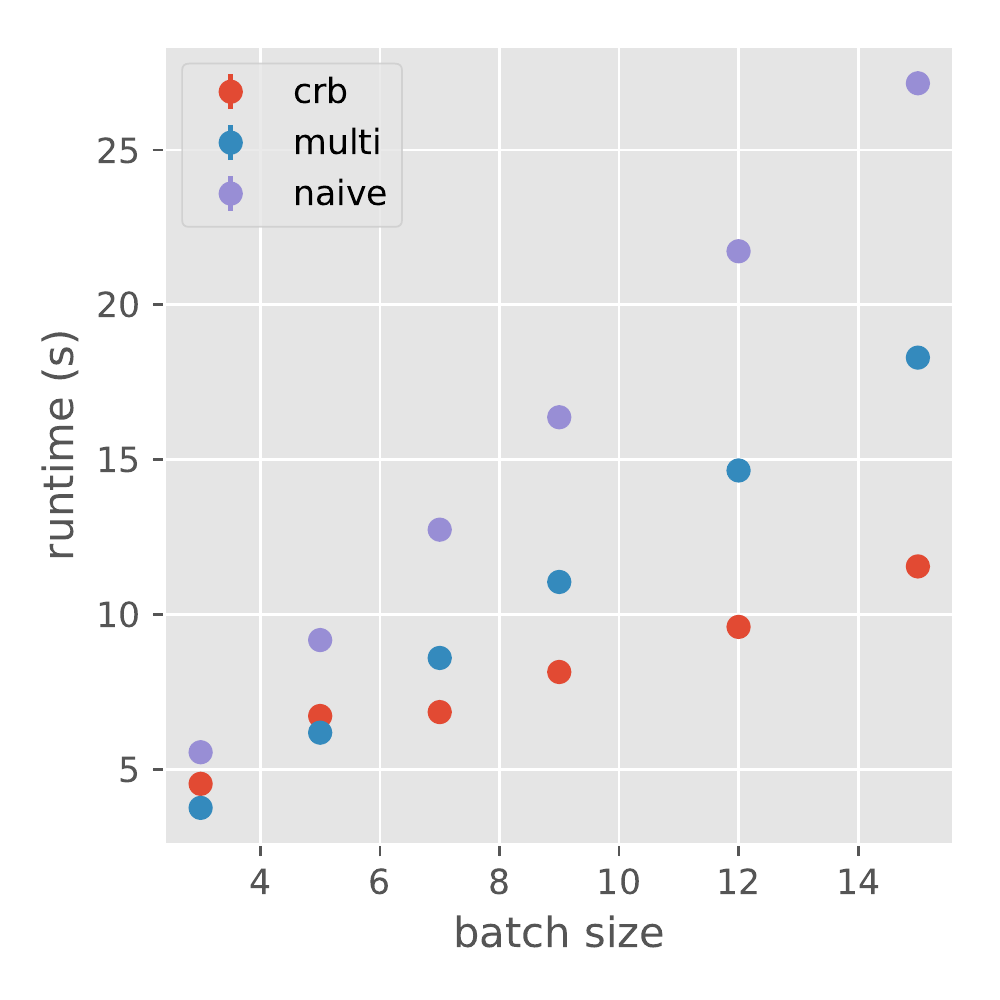}
    \caption{Runtime in seconds for processing 20 batches of different sizes,
    using the different strategies. Settings are similar to those of the
    previous experiment, except that i) the first layer now has 256 channels
    and ii) kernel size is set to 5, instead of 3. The channel rate is set to
    1, and the number of layers to 3.}
    \label{fig:experiment_bs}
\end{figure}

Depth and number of channels are not the only quantities that affect runtime. In
another experiment, we study how the runtime changes with batch size. For larger
batches, \texttt{crb} seems to be the method of choice. As shown in Fig.~
\ref{fig:experiment_bs}, both \texttt{naive} and \texttt{multi} lead to
a runtime which is linear over batch size; \texttt{crb}, however, seems to
be \emph{piecewise linear}: as the batch size increases, the slope
decreases. This behaviour is due, presumably, to the way \texttt{crb} is
able to exploit the GPU, transforming the original computation into a series
of new convolutions of different complexity.

\begin{figure}[ht]
    \centering
    \includegraphics[width=\textwidth]{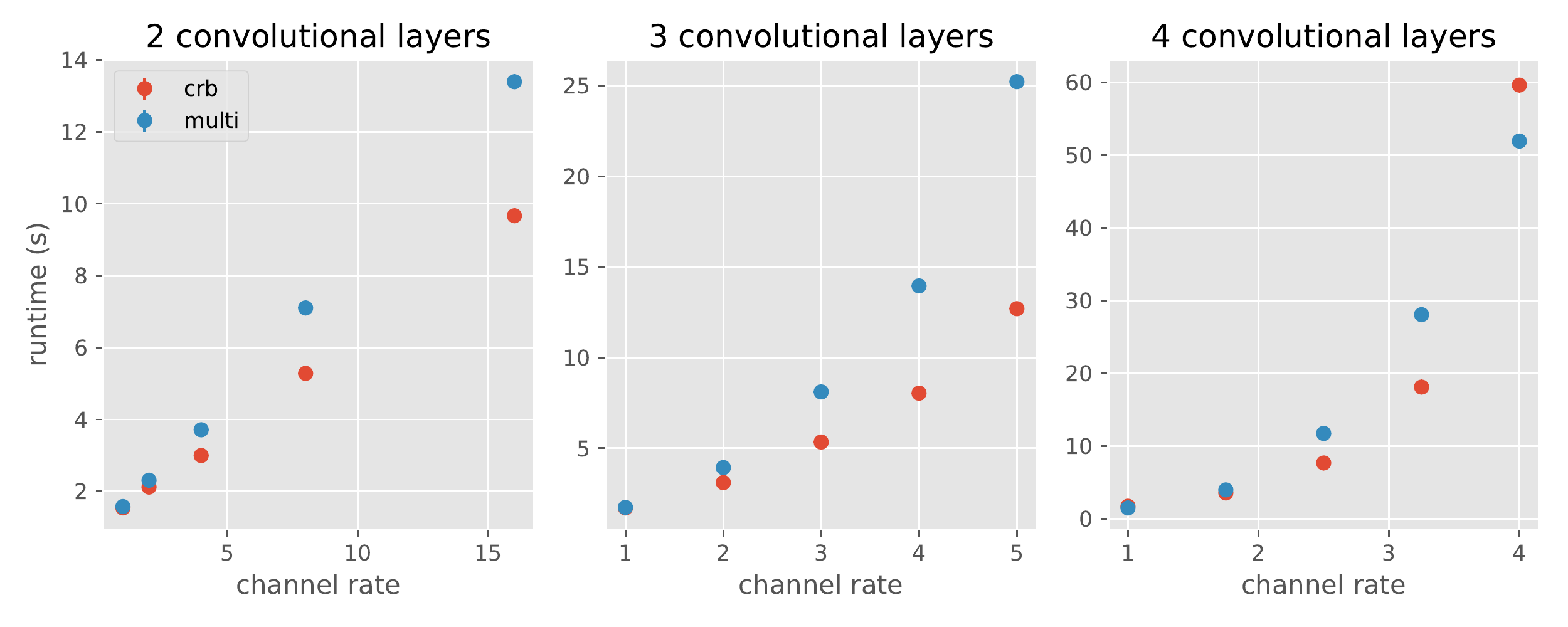}
    \caption{Analogous to Figure~\ref{fig:experiment_cnns}, except that kernel size of
    convolutions is set to 5 instead of 3. Increasing the kernel size seems to be an
    advantage to the \texttt{crb} strategy.}
    \label{fig:experiment_cnns2}
\end{figure}

Other factors can come into play, and it is not always intuitive to understand
which ones and why. For instance, increasing the kernel size of the convolutions
seems to be an advantage for \texttt{crb}, see Fig.~\ref{fig:experiment_cnns2}.

\subsection{Realistic networks}

At this point, one might wonder which of these approaches are better suited for
calculating per-example gradients for practical DCNNs. 
Such DCNNs typically contain many more than 4 layers, as well as widely varying
channel rates. To answer this question, we ran the same experiment for two
popular DCNN architectures, AlexNet
and VGG16. Runtime results are presented in Table~\ref{fig:experiments_compar}.

\begin{table}[ht]
    \begin{center}
        \begin{tabular}{cccccc} \toprule
            {Model} & {Batch Size} & {No DP (sec)} & {Naive (sec)} & {\texttt{crb} (sec)} & {\texttt{multi} (sec)} \\ \midrule
        {AlexNet} & 16 & $0.747 \pm 0.003$ & $30.17 \pm 0.02$ & $2.030 \pm 0.002$ & $3.076 \pm 0.005$ \\
        {VGG16} & 8 & $1.947 \pm 0.001$ & $37.20 \pm 0.10$ & $5.591 \pm 0.006$ & $4.630 \pm 0.020$ \\
            \bottomrule
        \end{tabular}
    \end{center}
    \caption{Runtime in seconds for processing 20 batches on AlexNet and VGG16.
    Inputs are randomly generated and have size $3 \times 256 \times 256$.
    For each model, we used a batch size that could reasonably fit into GPU
    memory. Implementations were taken from the \texttt{torchvision} library.}
    \label{fig:experiments_compar}
\end{table}

For relatively small networks such as AlexNet, \texttt{crb} performs up to
fifteen times times faster than \texttt{naive} on a Nvidia P100 GPU. It is also
slightly faster than \texttt{multi}. However, when
looking at the larger VGG16, \texttt{crb} becomes slightly slower than \texttt{multi}.
One could thus hypothesize that \texttt{multi} is the best option for larger networks;
as noted before, however, it is not obvious whether width and depth are the only relevant
quantities in play. Batch size, as well as the kernel size of the convolution,
seem to be of relevance as well.

Notice that we have not used batch normalization layers in any of the networks,
as they mix different examples on the batch and thus make per-gradient computations
impossible. For this same reason we have not tested CNNs which naturally include
batch normalization layers, such as the ResNet. An alternative is to use
instance normalization in cases when per-example gradient clipping is necessary.

\vskip1ex

Our experiments show that both \texttt{multi} and \texttt{crb} methods have
configurations in which they are the most efficient. Both methods have their
merits while using similar amount of GPU memory.

\section{Conclusion}

Both the existing \texttt{multi} approach and our extended \texttt{crb} are
capable of fully utilizing GPU capabilities to efficiently compute
per-example gradients. We have shown, empirically, that each is faster in a
particular region of the parameter space of DCNN architectures. 
In general, it is unclear which method will be more efficient.

Our approach is more complicated to be put in practice: it requires one to
adapt backpropagation hooks, as opposed to \texttt{multi}, for which
multiple copies of the model can be created on a higher level. Notice also
that our approach uses PyTorch's peculiarities---namely the \texttt{group}
argument in the convolutional layer---and should be adapted to different 
deep learning frameworks. Our hope is that the findings we present in
this work will be useful in furthering the development and improvement
of ML components necessary for privacy-aware machine learning.

We have implemented a PyTorch version extending \texttt{torch.nn}, which is
available at \url{https://github.com/owkin/grad-cnns}.

\printbibliography

\end{document}